\pdfoutput=1

\documentclass[11pt]{article}

\usepackage[preprint]{acl}

\usepackage{times}
\usepackage{latexsym}
\usepackage[T1]{fontenc}

\usepackage[utf8]{inputenc}

\usepackage{microtype}

\usepackage{inconsolata}

\usepackage{mathrsfs}
\usepackage{paralist}
\usepackage{booktabs}
\usepackage{multirow}
\usepackage{pgfplots}
\usepackage{scalefnt}
\usepackage{amsthm,amsmath,amssymb}
\usepackage{mathtools}
\usepackage{bm}
\usepackage{colortbl}
\usepackage{xcolor}
\usepackage{array}
\usepackage{stfloats}
\usepackage{tikz-dependency}
\usepackage{tikz}
\usepackage{siunitx}
\usepackage{graphicx}
\usepackage[ruled]{algorithm2e}
\usepackage{lineno}
\usepackage{subfigure}
\usepackage{helvet}
\usepackage{courier}
\usepackage{CJKutf8}
\definecolor{olivegreen}{rgb}{0.2,0.5,0.2}
\title{Ontology Memory-Augmented ASR Correction for Long Text-Speech Interleaved Conversations}

\author{
Xinxin Li$^{1,2,*}$, Huiyao Chen$^{1,2,}$\thanks{~~Authors contributed equally.}, Meishan Zhang$^{1,2}$\thanks{~~Corresponding author.}, Yunxin Li$^{1,2}$, \\ 
\textbf{Zulong Chen, Zhibo Ren, Xiaoqing Dong, Baotian Hu$^{1,2}$, Min Zhang$^{1,2}$} \\
$^{1}$Institute of Computing and Intelligence, Harbin Institute of Technology (Shenzhen), China \\
$^{2}$Shenzhen Loop Area Institute (SLAI), China \\
\texttt{lixinx0714@gmail.com, chenhy1018@gmail.com} \\
\texttt{mason.zms@gmail.com, zhangmin2021@hit.edu.cn}
}

\begin{document}
\begin{CJK}{UTF8}{gbsn}
\maketitle
\begin{abstract}
Automatic speech recognition (ASR) correction has traditionally focused on isolated utterances or short local contexts.
However, as text and speech become increasingly interleaved in long interactions, ASR correction requires conversation-level contextual evidence.
Existing ASR correction methods often rely on the current hypothesis or concatenate raw dialogue history.
In such contexts, sparse correction evidence can be difficult to locate amid redundancy and noise.
Addressing these challenges, we propose an ontology memory-augmented ASR correction framework for long text-speech interleaved conversations.
The framework organizes preceding interaction history into a dynamically updatable ontology memory, where entities, terminology, surface variants, potential ASR confusions, and semantic relations are stored as retrievable nodes for context-grounded correction.
To evaluate this setting, we construct RAMC-Corr, a dataset derived from MAGIC-RAMC for long-range ASR correction with grounded context.
Experiments on RAMC-Corr show that our method improves over direct correction in 9 out of 10 paired backbone-setting combinations and encourages more selective and evidence-grounded corrections for context-dependent ASR errors.
\end{abstract}

\setlength{\abovedisplayskip}{5pt}
\setlength{\belowdisplayskip}{5pt}
\setlength{\abovedisplayshortskip}{5pt}
\setlength{\belowdisplayshortskip}{5pt}

\section{Introduction}

Automatic speech recognition (ASR) serves as a fundamental component in a wide range of real-world applications, including voice assistants, meeting transcription, medical consultation, instant messaging, and intelligent office systems \cite{kim-etal-2025-voice-assistant,billings-mcdonnell-2025-connecting, hangchen-etal-2025-misp}.
With the rapid development of speech interaction, large language models, and multimodal assistants, ASR outputs are no longer used merely as isolated transcriptions of short voice commands.
Instead, they are increasingly embedded in continuous human-computer interaction, where text and speech appear alternately and jointly constitute the evolving interaction state of the current conversation \cite{zhang-etal-2023-speechgpt, DBLP:journals/corr/abs-2306-12925}.
Before a new speech segment arrives, the system may have already accumulated substantial preceding context, such as user-typed messages, confirmed dialogue history, meeting agendas, medical descriptions, or task instructions.
Such contextual information can provide valuable evidence for recognizing domain-specific entities, long-tail terminology, and semantically dependent expressions in subsequent speech segments, but how to effectively exploit it for ASR correction remains underexplored.

\begin{figure}[t]
    \centering
    \includegraphics[width=1.0\columnwidth]{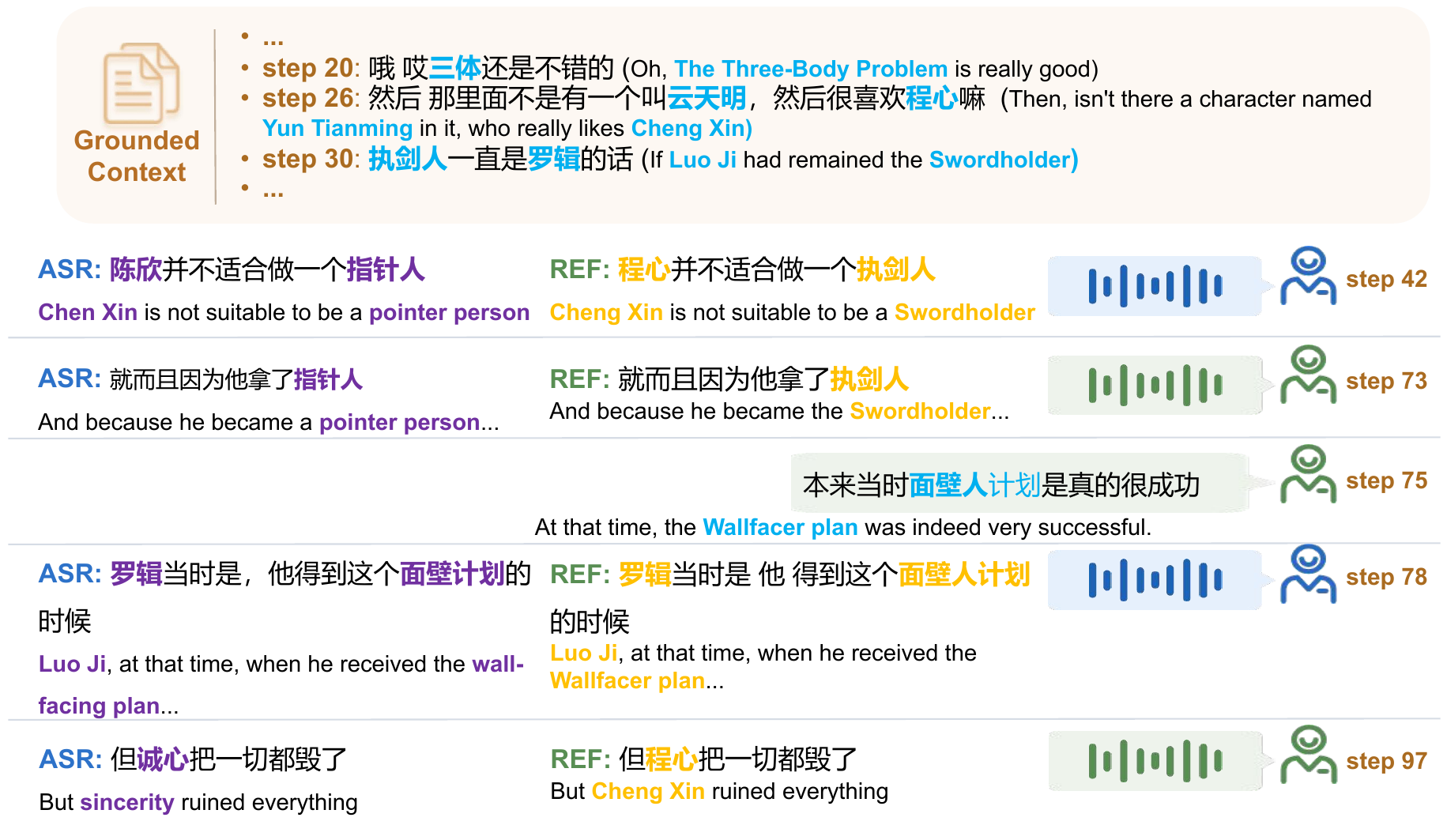}
    \caption{An example of context-grounded ASR correction in a long-range Chinese text-speech interleaved conversation. The original Chinese utterances are shown with English translations for readability. Earlier context in the same conversation provides evidence for correcting later ASR errors involving entities and domain-specific terms.}
    \label{fig:example}
\end{figure}

Figure~\ref{fig:example} shows a representative context-grounded ASR correction case from our Chinese dataset.
The grounded context introduces dialogue-specific entities and terms, including ``三体'', ``云天明'', ``程心'', ``罗辑'', and ``执剑人'', which later serve as evidence for correcting spoken ASR hypotheses.
For example, the ASR system misrecognizes the domain-specific term ``执剑人'' as the phonetically plausible but contextually incorrect ``指针人''.
This error is difficult to resolve from the local hypothesis alone, but can be corrected by grounding the correction in earlier conversational context.


These observations highlight the need for ASR correction methods that can exploit contextual evidence accumulated during an ongoing interaction.
Existing ASR correction methods typically take the current ASR hypothesis as input and use language models to rewrite erroneous transcriptions~\cite{DBLP:conf/asru/YangGLGBS23,DBLP:conf/nips/0075HYSCS23}.
Some studies further incorporate external vocabularies, entity lists, candidate phrases, or domain knowledge to improve the correction of named entities and rare words~\cite{ghosh-etal-2025-failing,im-etal-2025-deragec}.
However, these resources are usually static, whereas useful correction evidence in long text-speech interleaved conversations is often introduced dynamically by the interaction itself.
Simply concatenating historical context with the current ASR hypothesis is also insufficient, since relevant evidence becomes sparse and difficult to locate as the interaction grows longer.
Redundant turns, irrelevant information, and noisy ASR outputs can further obscure the contextual cues needed for correction.
These limitations motivate a structured memory mechanism that transforms preceding interaction context into reusable correction evidence.

In this work, we study \textit{context-grounded ASR correction} for long-range text-speech interleaved conversations.
The task requires the system to correct subsequent ASR transcriptions using causally accessible interaction history.
Different from utterance-level correction, the key challenge lies in how to organize, retrieve, and reuse previously established contextual knowledge, such as entities, aliases, domain concepts, and semantic constraints.

To address this challenge, we propose an ontology memory-augmented ASR correction framework.
The framework extracts correction-relevant knowledge from preceding interaction history and organizes it into a conversation-level ontology memory.
When a new ASR hypothesis arrives, the correction model retrieves relevant evidence from the ontology memory for context-grounded correction.
The memory is also dynamically updated as the interaction proceeds, enabling newly introduced concepts and terminology to support subsequent correction.

To evaluate this setting, we construct RAMC-Corr, a context-grounded ASR correction dataset based on the MagicData-RAMC speech chat corpus \cite{DBLP:conf/interspeech/YangCLYYCXJZZX022}.
RAMC-Corr evaluates whether a system can use causally accessible grounded context to correct later ASR hypotheses in long-range interactive scenarios.
Experiments on RAMC-Corr show that our method improves over direct correction in 9 out of 10 paired settings, demonstrating the effectiveness of explicit memory for context-grounded ASR correction.

In summary, our main contributions are as follows:
\begin{compactitem}
    \item We study context-grounded ASR correction for long-range text-speech interleaved conversations.
    
    \item We propose an ontology memory-augmented correction framework that organizes conversation-level contextual knowledge into a retrievable and dynamically updatable structured memory.
    
    \item We construct RAMC-Corr, a MagicData-RAMC-based dataset for evaluating long-range context-grounded ASR correction in text-speech interleaved conversations.
    
    \item Experiments across multiple LLM backbones show consistent gains over direct correction.
\end{compactitem}
The code and dataset are publicly available at \href{https://github.com/fangfang123gh/ontology-asr-correction}{github/fangfang123gh/ontology-asr-correction}.

\section{RAMC-Corr Dataset Construction}
We build RAMC-Corr, a new context-grounded ASR correction benchmark derived from MagicData-RAMC \cite{DBLP:conf/interspeech/YangCLYYCXJZZX022}.
It is designed to simulate long-range text-speech interleaved dialogue scenarios, where subsequent ASR hypotheses are corrected using causally accessible reliable textual history.
MagicData-RAMC is a Chinese multi-turn conversational speech corpus collected from real interaction scenarios, containing long temporally ordered conversations with speech segments and corresponding human transcriptions.
This structure makes it suitable for constructing long-range context-grounded ASR correction examples.

For each dialogue, we automatically select a boundary based on the human transcriptions.
The contiguous transcription segments from the beginning of the dialogue to the boundary are used as the grounded context, which serves as causally accessible textual evidence for subsequent ASR correction.
The segments after the boundary form the online target region, where text segments are directly observed and speech segments are represented by ASR hypotheses to be corrected.
The corresponding human transcriptions of these speech segments are used as references for evaluation.

To ensure that the grounded context provides informative guidance for later correction, we require semantic continuity across the selected boundary.
Specifically, we select candidate boundaries according to the contextual similarity between the grounded context and the initial part of the online target region, and filter out dialogues with weak cross-boundary semantic relatedness.
For each valid dialogue, we generate at most one sample to avoid scattered or highly overlapping correction instances from the same conversation.
The dataset statistics are summarized in Table~\ref{tab:data_statistic}, and the detailed construction procedure is presented in Appendix~\ref{sec:data_construction}.

\begin{table}[t]
\centering
\setlength{\tabcolsep}{10pt}
\resizebox{0.95\columnwidth}{!}{%
\begin{tabular}{lccccc}
\hline
\textbf{Split} & \#Conv. & \#Samples & \textbf{Boundary Ratio} & \textbf{Context Sim.} \\
\hline
Train & 268 & 268 & 0.076 & 0.654 \\
Dev   & 17  & 17  & 0.071 & 0.638 \\
Test  & 43  & 43  & 0.117 & 0.635 \\
\hline
All   & 328 & 328 & 0.081 & 0.650 \\
\hline
\end{tabular}%
}
\caption{Statistics of the RAMC-Corr dataset across training, development, and test splits. \#Conv is the number of conversations, \#Samples is the number of correction conversations, and Context Sim. is the contextual similarity used for boundary selection.}
\label{tab:data_statistic}
\end{table}

\section{Method}
\begin{figure*}[t]
    \centering
    \includegraphics[width=2.0\columnwidth]{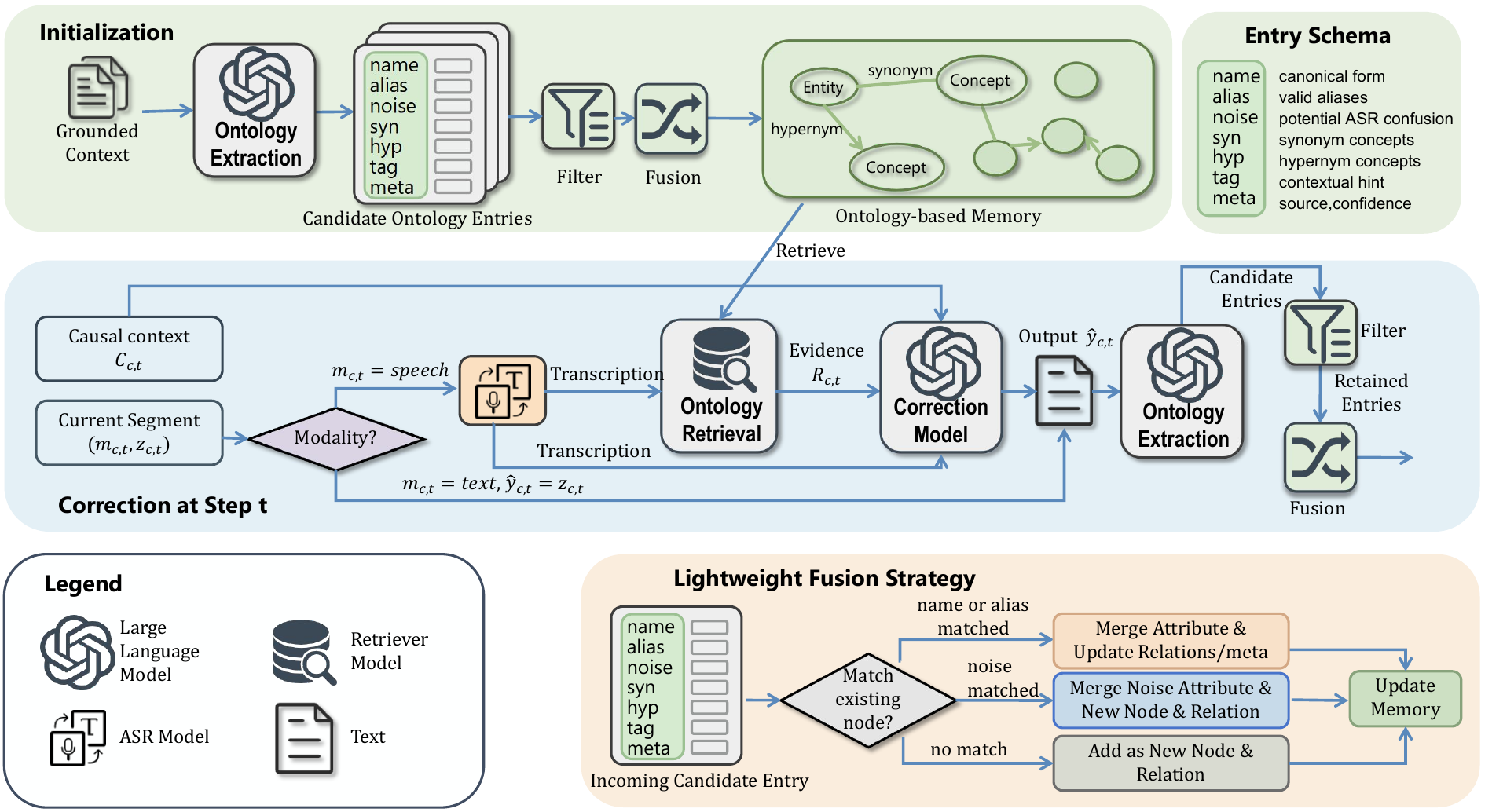}
    \caption{Overall architecture of the proposed ontology memory-augmented context-grounded ASR correction framework. }
    \label{fig:model}
\end{figure*}
A direct strategy is to feed the ASR hypothesis together with accessible dialogue history into a correction model.
However, this requires the model to locate useful entities, terminology, and semantic cues from an increasingly long and noisy context.

To address this issue, we propose an \textbf{ontology memory-augmented context-grounded ASR correction framework}, as illustrated in Fig.~\ref{fig:model}.
Instead of relying solely on raw dialogue history, the framework maintains a conversation-level ontology memory that incrementally organizes reusable entities, terminology, domain concepts, surface variants, and semantic relations.
The memory is first initialized from the grounded context and is then updated online after each processed segment.

During online processing, text segments are treated as reliable inputs and directly forwarded as outputs, while also being used to enrich the ontology memory.
For each speech segment, the base ASR hypothesis is corrected by a correction model conditioned on the causal context and relevant ontology evidence retrieved from the current memory.
After the output is produced, candidate ontology entries are extracted, filtered, and incorporated into the memory through a lightweight fusion strategy.

\subsection{Task Definition}

Given the $c$-th conversation sample, we consider a grounded context $\mathcal{G}_c$ followed by a temporally ordered processing sequence $\mathcal{X}_c$.
The grounded context contains information available before online processing begins, while the processing sequence consists of text and speech segments:
\begin{equation}
    \mathcal{X}_c = \{(m_{c,1}, z_{c,1}), \ldots, (m_{c,T_c}, z_{c,T_c})\},
\end{equation}
where $m_{c,t} \in \{\mathrm{text}, \mathrm{speech}\}$ denotes the modality of the $t$-th segment and $z_{c,t}$ denotes its content.
Text segments are regarded as reliable inputs, whereas speech segments are transcribed by a base ASR system into hypotheses $a_{c,t}$ and serve as the targets of correction.

Under the streaming setting, when processing step $t$, the system can only access the grounded context and previous outputs:
\begin{equation}
    C_{c,t} =
    \left(
    \mathcal{G}_c,
    \{\hat{y}_{c,i} \mid 1 \le i < t\}
    \right).
\end{equation}
This causal constraint prevents the system from using future segments.

The output at each step is defined as:
\begin{equation}
    \hat{y}_{c,t} =
    \begin{cases}
        z_{c,t},
        & m_{c,t} = \mathrm{text}, \\[4pt]
        F(a_{c,t}, C_{c,t}),
        & m_{c,t} = \mathrm{speech},
    \end{cases}
    \label{eq:objective}
\end{equation}
where $F(\cdot)$ denotes a correction function that refines the ASR hypothesis using only causally accessible information.

\subsection{Ontology Working Memory}

We maintain a conversation-level ontology working memory to transform long-range interaction history into reusable correction evidence.
Rather than a predefined static knowledge base, it is dynamically constructed within each conversation to store reusable entities, terminology, surface variants, and semantic relations.

We denote the memory before time step $t$ in conversation $c$ as $\mathcal{M}_{c,t}$, which consists of an ontology node set $\mathcal{V}_{c,t}$ and a semantic relation set $\mathcal{E}_{c,t}$.
Each node represents a reusable entity, item, or domain concept, together with its aliases, potential ASR noise forms, contextual tags, and metadata.
Relations encode semantic associations between concepts, such as synonym and hypernym relations.

\subsubsection{Candidate Ontology Entries}

Given causally accessible textual content, including the grounded context, reliable text segments, and corrected speech outputs, the ontology extractor produces a set of candidate ontology entries.
Each entry follows a lightweight schema with seven fields: \texttt{name}, \texttt{alias}, \texttt{noise}, \texttt{syn}, \texttt{hyp}, \texttt{tag}, and \texttt{meta}.
As shown in Figure~\ref{fig:model}, the schema records the canonical concept, aliases, potential ASR noise forms, synonym and hypernym links, contextual tags, and auxiliary metadata.
Candidate entries serve as the basic unit for memory writing.
During this process, the central concept and its variants are converted into ontology node attributes, while semantic associations such as synonymy and hypernymy are written into the relation set.

\subsubsection{Initialization and Lightweight Fusion}

Before online correction begins, we initialize the ontology working memory from the grounded context.
We view the grounded context $\mathcal{G}_c$ as an ordered sequence of context segments $\{g_{c,\ell}\}_{\ell=1}^{L_c}$.
Starting from an empty memory, the system processes these segments in order, extracts candidate ontology entries from each segment, and writes them into memory using the same update operation introduced above.
The resulting memory is used as the initial online memory $\mathcal{M}_{c,1}$.

Incoming entries are compared with existing nodes using the \texttt{name}, \texttt{alias}, and \texttt{noise} fields.
Matches on \texttt{name} or \texttt{alias} are treated as concept-level matches and are merged into the corresponding nodes.
Matches supported only by the \texttt{noise} field are used to expand ASR-confusion evidence, but do not by themselves prevent the incoming entry from being added as a new node.
For matched entries, node attributes and associated metadata are updated, while synonym and hypernym associations are incorporated into the relation set.
Entries without concept-level matches are added as new nodes.
When duplicated relations are observed, the system updates their metadata, such as occurrence count and confidence, instead of inserting redundant edges.
The same fusion strategy is reused during online processing.

\subsection{Ontology Evidence Retrieval}

Directly conditioning ASR correction on the entire interaction history is inefficient and prone to entity ambiguity.
We therefore retrieve structured ontology evidence from the current working memory, allowing the correction model to access localized contextual cues without searching through raw long-range history.

Because only speech segments require correction, ontology retrieval is performed only when the current segment is speech.
Given the ASR hypothesis $a_{c,t}$, causal context $C_{c,t}$, and current ontology memory $\mathcal{M}_{c,t}$, the retriever obtains relevant ontology evidence:
\begin{equation}
\begin{split}
R_{c,t}
=
\mathrm{Retrieve}_{\eta}
(&a_{c,t}, C_{c,t}, \mathcal{M}_{c,t}), \\
& m_{c,t} = \mathrm{speech}.
\end{split}
\end{equation}

The retrieved evidence $R_{c,t}$ consists of matched ontology nodes together with their stored attributes and associated relations, such as aliases, potential ASR confusion forms, contextual tags, and synonym or hypernym associations.
The retriever $\mathrm{Retrieve}_{\eta}$ is implementation-agnostic and can be instantiated with lexical matching, semantic retrieval, vector retrieval, or hybrid strategies.

\subsection{Ontology-Augmented ASR Correction and Memory Update}

\paragraph{ASR Correction.}
For a speech segment, the correction model generates the final transcription conditioned on the ASR hypothesis, causal context, and retrieved ontology evidence.
We instantiate the correction function $F$ in Eq.~\eqref{eq:objective} as a parameterized model $f_{\theta}$:
\begin{equation}
    \hat{y}_{c,t}
    =
    f_{\theta}
    \left(
    a_{c,t},
    C_{c,t},
    R_{c,t}
    \right),
    \quad
    m_{c,t} = \mathrm{speech}.
    \label{eq:correction}
\end{equation}
The retrieved ontology evidence provides localized concept constraints and semantic cues that help resolve entity ambiguity and ASR confusions.
For text segments, no correction is applied, and the reliable text is directly used as the output:
\begin{equation}
    \hat{y}_{c,t} = z_{c,t},
    \quad
    m_{c,t} = \mathrm{text}.
\end{equation}

\paragraph{Memory Update.}
After the output $\hat{y}_{c,t}$ is determined, both corrected speech and reliable text are used as sources for memory enrichment.
The ontology extractor obtains candidate entries from the current output:
\begin{equation}
    \mathcal{B}_{c,t}
    =
    E_{\phi}
    \left(
    \hat{y}_{c,t}
    \right).
\end{equation}
Unsupported, unstable, or non-reusable entries are filtered out, yielding the retained subset $\mathcal{B}_{c,t}^{+}$.
The retained entries are then incorporated into the ontology memory using the lightweight fusion strategy described above:
\begin{equation}
    \mathcal{M}_{c,t+1}
    =
    \mathrm{Update}
    \left(
    \mathcal{M}_{c,t},
    \mathcal{B}_{c,t}^{+}
    \right).
\end{equation}
If no valid entry is retained, the memory remains unchanged.

\section{Experiments}
\subsection{Settings}



\paragraph{Evaluation Metrics.}
We report character error rate (CER) in two forms: corpus CER (C-CER) and macro CER (M-CER), along with relative CER reduction (RelCER).
C-CER aggregates edit distances over the whole set, while M-CER averages CER over individual correction instances.
CER is computed with jiwer\footnote{\url{https://github.com/jitsi/jiwer}} after NFKC normalization, lowercasing, and removing punctuation and whitespace.
RelCER is computed against the original ASR hypotheses:
$$
\mathrm{RelCER} =
\frac{
\mathrm{C\text{-}CER}_{\mathrm{Raw}} -
\mathrm{C\text{-}CER}_{\mathrm{Corrected}}
}{
\mathrm{C\text{-}CER}_{\mathrm{Raw}}
}
\times 100.
$$
Lower C-CER/M-CER and higher RelCER indicate better correction quality.
All values are percentages.

\paragraph{Model Details.}
We use \texttt{inf-retriever-v1} to retrieve ontology entries from working memory.
The ontology extractor and correction model use the same backbone LLM unless otherwise specified.
For correction, we evaluate Qwen2.5-7B/14B/72B-Instruct and Qwen3.5-4B/9B.
We also evaluate Gemma-4-26B-128K as a Full-History baseline, where the complete available history is directly provided without explicit working memory construction.
Fro asr model, we use Qwen2-Audio-7B-Instruct.
For readability, we omit the ``Instruct'' suffix in tables.
All LLM backbones are evaluated in a training-free inference-only setting.
We use vLLM \cite{DBLP:conf/sosp/KwonLZ0ZY0ZS23} for efficient model inference, and report all results from a single run unless otherwise specified.


\paragraph{Computational Setup.}
All experiments are inference-only and do not involve model training.
Experiments were conducted on NVIDIA H100 SXM5 80GB GPUs.

\paragraph{Hyperparameters.}
Unless otherwise specified, we use 5 previous utterances as local context and retrieve the top 10 ontology entries from the working memory.
For model inference, we set the temperature to 0.0, top-p to 1.0, and the maximum number of generated tokens to 2048.
In the few-shot setting, both ontology information extraction and correction use 3-shot demonstrations as the default prompting hyperparameter.
The demonstrations are manually selected from the development set and fixed across test examples within the same experimental setting.
By default, all segments after the boundary are provided as speech inputs, with no text inputs given for the target region.
In the ablation studies, we randomly replace a specified proportion of target-region speech inputs with their transcript form according to the predefined text-input ratio.
We further analyze the effect of the number of in-context examples in Appendix~\ref{app:example_ablation}, and provide the prompt templates in Appendix~\ref{app:prompt_templates}.

\subsection{Main Results}
\begin{table}[t]
    \centering
    \resizebox{0.98\columnwidth}{!}{
        \begin{tabular}{lllccc}
            \toprule
            \textbf{Model} & \textbf{Setting} & \textbf{Method} 
            & \textbf{C-CER $\downarrow$} 
            & \textbf{M-CER $\downarrow$} 
            & \textbf{RelCER $\uparrow$} \\
            \midrule
            Raw ASR & -- & -- & 27.94 & 55.44 & 0.00 \\
            \midrule

            \multirow{2}{*}{Gemma-4-26B}
            & ZS & Full-History & 27.94 & 55.44 & +0.50 \\
            & FS & Full-History & 27.96 & 55.44 & 0.00 \\
            \midrule

            \multirow{4}{*}{Qwen2.5-7B}
            & \multirow{2}{*}{ZS}
            & Direct & 36.40 & 73.01 & -30.27 \\
            & & Ours   & \textbf{31.76} {\color{olivegreen}(+12.7)} & 58.33 {\color{olivegreen}(+20.1)} & -13.67 \\
            & \multirow{2}{*}{FS}
            & Direct & 35.50 & 67.59 & -27.05 \\
            & & Ours   & 33.00 {\color{olivegreen}(+7.0)} & 59.34 {\color{olivegreen}(+12.2)} & -18.10 \\
            \midrule

            \multirow{4}{*}{Qwen2.5-14B}
            & \multirow{2}{*}{ZS}
            & Direct & 35.66 & 84.57 & -27.64 \\
            & & Ours   & \textbf{29.04} {\color{olivegreen}(+18.6)} & 55.35 {\color{olivegreen}(+34.6)} & -3.95 \\
            & \multirow{2}{*}{FS}
            & Direct & 34.60 & 77.40 & -23.85 \\
            & & Ours   & 32.43 {\color{olivegreen}(+6.3)} & 66.30 {\color{olivegreen}(+14.3)} & -16.09 \\
            \midrule

            \multirow{4}{*}{Qwen2.5-72B}
            & \multirow{2}{*}{ZS}
            & Direct & 29.82 & 57.19 & -6.73 \\
            & & Ours   & \textbf{29.16} {\color{olivegreen}(+2.2)} & 55.62 {\color{olivegreen}(+2.7)} & -4.36 \\
            & \multirow{2}{*}{FS}
            & Direct & 30.38 & 58.31 & -8.73 \\
            & & Ours   & 30.06 {\color{olivegreen}(+1.1)} & 57.85 {\color{olivegreen}(+0.8)} & -7.58 \\
            \midrule

            \multirow{4}{*}{Qwen3.5-4B}
            & \multirow{2}{*}{ZS}
            & \cellcolor{blue!15}Direct & \cellcolor{blue!15}27.61 & \cellcolor{blue!15}55.17 & \cellcolor{blue!15}+1.19 \\
            & & \cellcolor{blue!15}Ours   & \cellcolor{blue!15}27.36 {\color{olivegreen}(+0.9)} & \cellcolor{blue!15}54.94 {\color{olivegreen}(+0.4)} & \cellcolor{blue!15}+2.08 \\
            & \multirow{2}{*}{FS}
            & Direct & 29.16 & 57.60 & -4.36 \\
            & & \cellcolor{blue!15}Ours   & \cellcolor{blue!15}\textbf{27.32} {\color{olivegreen}(+6.3)} & \cellcolor{blue!15}54.97 {\color{olivegreen}(+4.6)} & \cellcolor{blue!15}+2.20 \\
            \midrule

            \multirow{4}{*}{Qwen3.5-9B}
            & \multirow{2}{*}{ZS}
            & \cellcolor{blue!15}Direct & \cellcolor{blue!15}27.47 & \cellcolor{blue!15}55.29 & \cellcolor{blue!15}+1.68 \\
            & & \cellcolor{blue!15}Ours   & \cellcolor{blue!15}\textbf{27.26} {\color{olivegreen}(+0.8)} & \cellcolor{blue!15}55.17 {\color{olivegreen}(+0.2)} & \cellcolor{blue!15}+2.42 \\
            & \multirow{2}{*}{FS}
            & Direct & 28.02 & 55.85 & -0.30 \\
            & & Ours   & 28.89 {\color{red}(-3.1)} & 58.05 {\color{red}(-3.9)} & -3.41 \\
            \bottomrule

        \end{tabular}
    }
    \caption{Main results on the RAMC-Corr dataset.
All values are percentages, and ZS/FS denote zero-shot/few-shot prompting.
Values in parentheses indicate the relative reduction over Direct for C-CER and M-CER under the same backbone and prompting setting.
Bold numbers indicate the lower C-CER between Direct and Ours.}
    \label{tab:main_result}
\end{table}

Table~\ref{tab:main_result} reports the main correction results on the RAMC-Corr dataset.
We compare Direct correction, which feeds ASR hypotheses and recent corrected conversational context directly to the backbone model, with our memory-augmented correction method, which uses an explicit working memory constructed from dialogue history.
We also include a Full-History baseline using Gemma-4-26B-128K, where the complete available conversation history is provided to the model without explicit memory construction.
Overall, our method improves C-CER over Direct in 9 out of 10 paired backbone-setting combinations, indicating that explicit working memory provides useful correction evidence beyond local prompting.

Direct correction is often unstable for ASR correction.
In several settings, it produces higher CER than the original ASR hypotheses, suggesting that unconstrained LLM rewriting can introduce harmful edits.
Our method mitigates this issue by grounding corrections in retrieved memory and structured dialogue evidence.
For example, under zero-shot prompting, our method reduces C-CER from 36.40 to 31.76 with Qwen2.5-7B and from 35.66 to 29.04 with Qwen2.5-14B, corresponding to 12.7\% and 18.6\% relative reductions, respectively.
For Qwen2.5-14B under zero-shot prompting, our method reduces C-CER from 35.66 to 29.04 and M-CER from 84.57 to 55.35, corresponding to relative reductions of 18.6\% and 34.6\%, respectively.
Our method also improves cases where Direct is already competitive, such as the Qwen3.5-4B few-shot setting, where RelCER changes from -4.36\% to 2.20\%.

The Full-History baseline shows that simply exposing a long-context model to more historical text does not necessarily improve correction, as Gemma-4-26B-128K remains close to Raw ASR in both prompting settings.
The gains are not uniform across all settings.
Our method underperforms Direct in the Qwen3.5-9B few-shot setting, despite improving over Direct for the same backbone under zero-shot prompting.
This indicates that few-shot demonstrations may interact with memory guidance and sometimes encourage unnecessary modifications under an inference-only setting.
We therefore analyze correction behavior below to distinguish targeted improvements from harmful edits.

\subsection{Analysis}

\begin{figure}[t]
    \centering
    \includegraphics[width=1.00\linewidth]{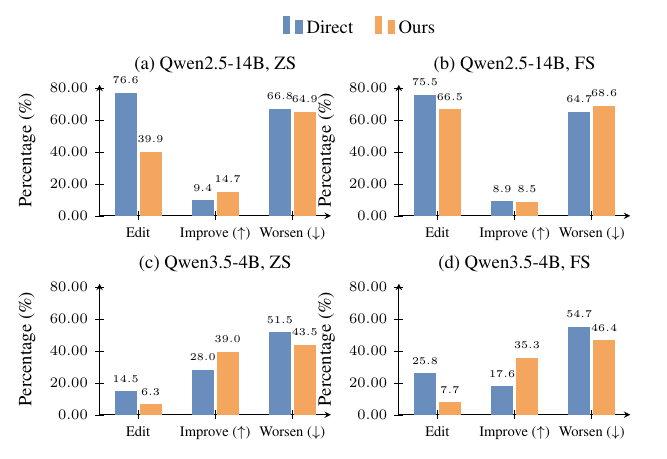}
    \caption{Edit behavior comparison under zero-shot and few-shot settings. 
Edit, Improve, and Worsen denote Edit Rate, Improve@Edit, and Worsen@Edit, respectively. ZS/FS denote zero-shot/few-shot prompting.}
    \label{fig:edit_performance}
\end{figure}

\paragraph{Correction behavior analysis.}

CER measures final quality but hides how a model edits ASR hypotheses.
We therefore use Edit Rate, Improve@Edit, and Worsen@Edit to measure whether edits occur, reduce CER, or increase CER.
Figure~\ref{fig:edit_performance} compares two representative backbones.
Qwen2.5-14B illustrates a high over-editing case, while Qwen3.5-4B represents a competitive setting where memory improves edit quality.

For Qwen2.5-14B, ontology memory mainly reduces excessive rewriting, although Worsen@Edit remains high after the model chooses to edit.
For Qwen3.5-4B, ontology memory shows a more consistent pattern by reducing Edit Rate, increasing Improve@Edit, and decreasing Worsen@Edit in both prompting settings.
These results suggest that ontology memory mainly serves as an edit-selection mechanism, while its effect on edit precision depends on how effectively the correction model uses the retrieved evidence.
Appendix~\ref{app:evidence_utilization} further analyzes how retrieval availability is associated with edit outcomes.


\paragraph{Impact of text-speech interleaving ratio.}

\begin{figure}[t]
    \centering
    \includegraphics[width=1.00\linewidth]{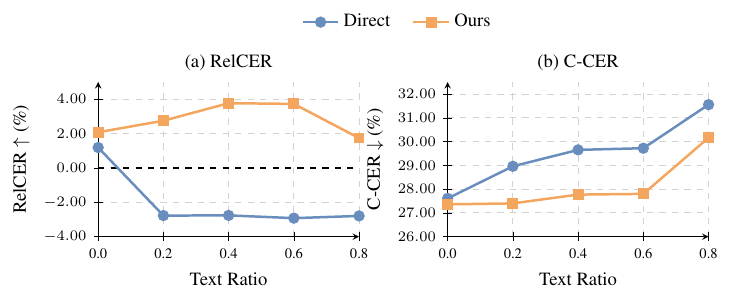}
    \caption{Impact of the text-input ratio on ASR correction performance. }
    \label{fig:portion_impact}
\end{figure}
We study whether RAMC-Corr remains effective when subsequent utterances are provided in different modalities.
Specifically, we vary the proportion of subsequent utterances represented as text transcriptions from 0.0 to 0.8, where 0.0 denotes the speech-only setting.
Experiments are conducted with Qwen3.5-4B under the zero-shot setting.
As shown in Figure~\ref{fig:portion_impact}, our method consistently achieves better RelCER and C-CER than Direct across all text ratios.
The improvement becomes more pronounced after text transcriptions are introduced, especially at moderate ratios of 0.4 and 0.6.
Direct, however, becomes less reliable under mixed text-speech inputs, leading to degraded RelCER.
When the ratio increases to 0.8, the gain becomes smaller, indicating that more text context is not always more useful unless it provides correction-relevant cues.
These results demonstrate the robustness of ontology memory under text-speech interleaved inputs.

\paragraph{Online latency in continuous speech interaction.}
\begin{table}[t]
\centering
\small
\resizebox{\columnwidth}{!}{
\begin{tabular}{lcccc}
\toprule
\multirow{2}{*}{\textbf{Metric}} &
\multirow{2}{*}{\textbf{Direct}} &
\multicolumn{3}{c}{\textbf{Ours}} \\
\cmidrule(lr){3-5}
&
&
\textbf{Top-k=5} &
\textbf{Top-k=10} &
\textbf{Top-k=15} \\
\midrule
Subseq. Speech Window
& 4.59 s
& 4.59 s
& 4.59 s
& 4.59 s \\

User Wait Mean / P95
& 0.01 / 0.00 s
& 0.92 / 3.67 s
& 0.97 / 3.76 s
& 0.99 / 3.85 s \\

Visible Corr. Mean / P95
& 0.53 / 0.90 s
& 2.31 / 4.39 s
& 2.35 / 4.51 s
& 2.38 / 4.51 s \\

Backend Mean
& 0.53 s
& 2.98 s
& 3.04 s
& 3.08 s \\

RelCER $\uparrow$
& 1.19
& 1.37
& 2.08
& 2.05 \\

\bottomrule
\end{tabular}
}
\caption{Online latency analysis under the Qwen3.5-4B zero-shot setting with continuous speech input.Ours is evaluated with different ontology retrieval sizes.}
\label{tab:online_latency_relcer}
\end{table}

We analyze latency in a continuous speech-input scenario, where backend computation for the previous chunk can overlap with the following speech segment and its ASR processing.
Since this overlap can hide part of the computation from the user, we report user wait at the chunk level, together with backend latency and visible correction latency.
Backend latency measures the end-to-end computation time for correction and memory operations, while visible correction latency measures the delay from ASR decoding to the corrected output.
Direct denotes correction without working memory, while Ours denotes correction with memory retrieval and ontology update.

As shown in Table~\ref{tab:online_latency_relcer}, Ours increases backend latency from 0.53 s to 2.98--3.08 s.
However, the average user wait remains below 1 s across all Top-k settings because part of the computation is hidden by the 4.59 s speech-input window.
Increasing Top-k from 5 to 15 only slightly increases average user wait from 0.92 s to 0.99 s, suggesting that larger retrieval sizes have limited impact on user-perceived latency in this setting.


\paragraph{Case study.}
\begin{table}[t]
    \centering

\begin{tikzpicture}[node distance = 0cm, auto]
    \def\mytabletwo{
        \renewcommand{\arraystretch}{1.18}
        \resizebox{1.0\columnwidth}{!}{
        \begin{tabular}{
            cc
            >{\raggedright\arraybackslash}p{8.0cm}
            >{\raggedright\arraybackslash}p{4.0cm}
        }
            \toprule
            \textbf{Num} &  & \textbf{Outputs} & \textbf{Memory Evidence} \\
            \midrule

            \textbf{1} & \textbf{Before} 
            & 真的你在\textbf{院线汇}不管你是\textbf{院线汇}还是西 \newline
              \textit{Really you are in the \textbf{cinema line association} whether you are in the \textbf{cinema line association} or west}
            & \multirow{4}{4.0cm}{
              A retrieved memory about \textbf{Xiong Liang} (\textbf{熊亮}) mentions the graduate student union (\textbf{院学生会}), supporting the organization-name correction.

              } \\

            & \cellcolor{gray!15}\textbf{After Direct} 
            & \cellcolor{gray!15}真的你在\textbf{院线汇}不管你是\textbf{院线汇}还是西 \newline
              \cellcolor{gray!15}\textit{Really you are in the \textbf{cinema line association} whether you are in the \textbf{cinema line association} or west}
            & \\

            & \cellcolor{gray!15}\textbf{After Ours} 
            & \cellcolor{gray!15}真的你在\textbf{院学生会}不管你是\textbf{院学生会}还是西 \newline
              \cellcolor{gray!15}\textit{Really you are in the \textbf{graduate student union} whether you are in the \textbf{graduate student union} or west}
            & \\

            & \cellcolor{gray!15}\textbf{Reference} 
            & \cellcolor{gray!15}真的你在\textbf{院学生会}不管你是在\textbf{院学生会}还是 \newline
              \cellcolor{gray!15}\textit{Really you are in the \textbf{graduate student union} whether you are in the \textbf{graduate student union} or}
            & \\

            \addlinespace[3pt]

            \textbf{2} & \textbf{Before} 
            & 就是这个\textbf{悲伤有成河}吗 \newline
              \textit{Is it this \textbf{sadness becomes a river}}
            & \multirow{4}{4.0cm}{
          A retrieved memory about \textbf{Yi Yao} (\textbf{易遥}) links the dialogue context to the movie title \textit{The Disillusioned Youth} (\textbf{悲伤逆流成河}).

          } \\

            & \cellcolor{gray!15}\textbf{After Direct} 
            & \cellcolor{gray!15}就是这个\textbf{悲伤有长河}吗 \newline
              \cellcolor{gray!15}\textit{Is it this \textbf{sadness has a long river}}
            & \\

            & \cellcolor{gray!15}\textbf{After Ours} 
            & \cellcolor{gray!15}就是这个\textbf{悲伤逆流成河}吗 \newline
              \cellcolor{gray!15}\textit{Is it this \textbf{The Disillusioned Youth}}
            & \\

            & \cellcolor{gray!15}\textbf{Reference} 
            & \cellcolor{gray!15}就是这个\textbf{悲伤逆流成河}吗 \newline
              \cellcolor{gray!15}\textit{Is it this \textbf{The Disillusioned Youth}}
            & \\

            \addlinespace[3pt]

            \textbf{3} & \textbf{Before} 
            & 你看了是\textbf{十宗罪}呀 \newline
              \textit{What you watched was \textbf{The Ten Deadly Sins}}
            & \multirow{4}{4.0cm}{
              The retrieved memory for \textit{The Ten Deadly Sins} (\textbf{十宗罪}) records prior ASR noise forms such as ``missing crime'' (\textbf{失踪罪}) and ``Shi Zhongyu'' (\textbf{石中玉}).

              } \\

            & \cellcolor{gray!15}\textbf{After Direct} 
            & \cellcolor{gray!15}你看了是\textbf{失踪者}呀 \newline
              \cellcolor{gray!15}\textit{What you watched was \textbf{The Missing}}
            & \\

            & \cellcolor{gray!15}\textbf{After Ours} 
            & \cellcolor{gray!15}你看了是\textbf{十宗罪}呀 \newline
              \cellcolor{gray!15}\textit{What you watched was \textbf{The Ten Deadly Sins}}
            & \\

            & \cellcolor{gray!15}\textbf{Reference} 
            & \cellcolor{gray!15}你看的是\textbf{十宗罪}啊 \newline
              \cellcolor{gray!15}\textit{What you watched was \textbf{The Ten Deadly Sins}}
            & \\

            \bottomrule
        \end{tabular}}}

    \newlength\figtwoheight
    \newlength\figtwowidth

    \newlength\figtwotableinter
    \setlength\figtwotableinter{\heightof{\mytabletwo}}
    \addtolength{\figtwotableinter}{0.5\figtwoheight}

    \node[inner sep=0pt, node distance = \figtwotableinter] (tab) {\mytabletwo};

\end{tikzpicture}

    \caption{
    Qualitative examples of ontology-memory-augmented ASR correction.
    \textbf{Before} shows the original ASR transcription.
    \textbf{After Direct} shows correction without ontology memory.
    \textbf{After Ours} shows correction with ontology memory.
    \textbf{Memory Evidence} shows the retrieved memory evidence provided as contextual support.
    Bold text highlights the key corrected or over-corrected expressions.
    }
    \label{tab:case_study}
\end{table}
Table~\ref{tab:case_study} illustrates how ontology memory improves context-grounded ASR correction through retrieved evidence. 
Without such memory, the model tends to rely on linguistic plausibility, resulting in incomplete entity recovery or unsupported substitutions. 
For example, the Direcct fails to recover ``院学生会'' (graduate student union) from ``院线汇'' (cinema line association), and incorrectly rewrites the correct title ``十宗罪'' (The Ten Deadly Sins) as ``失踪者'' (The Missing). 
In contrast, our method grounds corrections in retrieved dialogue-relevant evidence, including organization context for ``院学生会'', movie-title context for ``悲伤逆流成河'' (The Disillusioned Youth), and prior noise-form evidence for preserving ``十宗罪''. 
These examples show that ontology memory supports more selective and contextually grounded edits rather than broader rewriting.

\section{Related Work}

Post-ASR correction improves transcription quality without modifying the underlying ASR system.
Early methods relied on language model rescoring, confusion modeling, and n-best hypothesis selection~\cite{DBLP:journals/corr/abs-2306-12925, DBLP:conf/interspeech/RunarsdottirHG19}.
With the development of large language models, recent studies have increasingly formulated ASR correction as a conditional generation task, where the model regenerates more accurate and fluent text conditioned on noisy ASR outputs \cite{DBLP:conf/asru/YangGLGBS23,DBLP:conf/nips/0075HYSCS23}.
Similar evidence has also been observed in structured language understanding, where self-correction help LLMs produce more consistent structured outputs~\cite{li-etal-2025-llms-also,DBLP:journals/corr/abs-2512-10004}.
Despite their strong performance, most generative ASR correction methods still focus on the current utterance or local segment, with limited modeling of contextual knowledge accumulated from prior interactions.

To improve the correction of named entities, rare words, and domain-specific terminology, another line of research introduces contextual biasing or retrieval-augmented mechanisms that incorporate external vocabularies, entity databases, candidate phrases, or domain knowledge as auxiliary evidence for ASR correction \cite{DBLP:conf/bigdataconf/ZhangQZSLLZPPTYJ23, DBLP:journals/corr/abs-2410-13198}. \citet{ghosh-etal-2025-failing} and \citet{im-etal-2025-deragec} further combine semantic and phonetic similarity to alleviate generalization issues caused by open-vocabulary settings and acoustic confusion. These studies demonstrate the importance of external contextual evidence for ASR correction. However, most existing approaches still rely on static knowledge sources or independently retrieve candidate items for each utterance. In contrast, correction evidence in long-form interleaved text-speech conversations often naturally originates from prior interactions themselves.

As interaction duration increases, these limitations become more pronounced.
Long-form conversational ASR must handle extended speech inputs, topic shifts, context truncation, cross-segment error propagation, and concept consistency, since ASR-induced errors may further propagate to downstream semantic analysis and degrade task performance \cite{chen-etal-2024-semantic, kang-etal-2025-llase}.
Recent studies have explored long-context speech understanding via benchmarks, speech compression, multimodal modeling, and long-context modeling \cite{DBLP:journals/corr/abs-2601-13539,DBLP:journals/corr/abs-2603-06193,DBLP:journals/corr/abs-2507-13264,DBLP:journals/taslp/WeiLLLJX24}.
While these studies improve long-audio modeling, cross-segment continuity, and multimodal understanding, they pay relatively little attention to how textual and situational information established during prior interactions can be organized into reusable evidence for subsequent ASR correction.


Meanwhile, memory-augmented language models have been widely studied in tasks such as personalized dialogue, long-context question answering, and information compression \cite{ke-etal-2025-flexibly,chen-etal-2025-compress}. However, memory mechanisms have rarely been explored for post-ASR correction, especially in interleaved text-speech conversations where interaction context must be continuously accumulated, structurally organized, and dynamically retrieved for future correction.

\section{Conclusion}

We studied ASR correction in text-speech interleaved conversations, where long-range interaction history can provide valuable but noisy correction evidence.
To exploit such context, we proposed an ontology-memory-augmented framework that organizes reliable dialogue history into a dynamically updatable working memory and retrieves structured evidence for subsequent ASR correction.
We constructed RAMC-Corr from MagicData-RAMC to evaluate long-range context-grounded correction.
Experiments across multiple LLM backbones show that our method improves over direct correction in 9 out of 10 paired settings.
Further analyses indicate that these gains mainly come from more selective and evidence-grounded edits rather than indiscriminate rewriting, showing the potential of ontology memory for interactive ASR correction.




\section*{Limitations}

Our work has four major limitations.
First, RAMC-Corr is currently constructed from Chinese text-speech interleaved conversations.
Although this setting provides a useful testbed for studying long-range context-grounded ASR correction, it does not fully cover multilingual, cross-lingual, or domain-diverse scenarios.
In the future, we plan to extend the benchmark to more languages and real-world application domains.

Second, our experiments mainly focus on a training-free LLM setting with lightweight few-shot prompting.
This design allows us to evaluate the plug-and-play capability of ontology memory across different LLM backbones.
However, the potential benefits of supervised fine-tuning, instruction tuning, parameter-efficient adaptation, or more advanced prompting strategies remain underexplored.
We leave a systematic investigation of these model adaptation strategies to future work.

Third, the retrieval component in our framework is not specifically optimized for ASR correction.
The current retrieval strategy may not fully capture correction-specific cues such as phonetic similarity, entity recurrence, acoustic confusions, and discourse dependencies.
Future work can develop task-aware retrievers that jointly consider semantic, phonetic, and conversational signals for more accurate context selection.

Finally, our evaluation is conducted on a benchmark derived from existing text-speech interleaved data.
Although RAMC-Corr supports controlled evaluation of long-range context use, real-world interactive speech applications may involve more diverse speakers, background noise, spontaneous speech, and changing dialogue states.
We plan to further validate the framework in more realistic and open-ended interaction scenarios.

\section*{Ethical Statement}

We construct RAMC-Corr based on MagicData-RAMC for long-range context-grounded ASR correction in text-speech interleaved conversations.
RAMC-Corr is derived from the existing MagicData-RAMC dataset and inherits its original speech data and reference transcripts.
We do not collect new speech recordings, textual content, personal information, or private user data during dataset construction.
The dataset is used only for research purposes, following the license and usage conditions of the original data source.

RAMC-Corr defines the evaluation split and correction targets for context-grounded ASR correction.
The ASR hypotheses used in our experiments are generated by ASR systems during the experimental pipeline and are not newly collected human data.
No personally identifiable information is introduced, and no additional sensitive or harmful content is added.
The dataset is intended for evaluating context-grounded ASR correction rather than identifying or profiling individual speakers.

We acknowledge that automatic ASR correction may produce incorrect or unsupported modifications.
Therefore, corrected transcripts should be used with caution in high-stakes scenarios, and human verification is recommended when they are used for consequential decisions.

Overall, the construction and use of RAMC-Corr are compliant with research ethics.

\bibliography{custom}

\appendix

\section{Appendix}
\label{sec:appendix}

\subsection{RAMC-Corr Dataset Construction Details}
\label{sec:data_construction}

This section provides additional details on the sample construction process of RAMC-Corr. Given an original conversation from MagicData-RAMC, we select a split boundary from its temporally ordered speech segments and manual transcriptions. The manual transcriptions before the boundary are used as the trusted prefix context, while the speech segments after the boundary form the target region to be corrected. The boundary selection process is based only on manual transcription texts, ensuring that the split does not depend on ASR outputs from the target region.

For a conversation containing $n$ segments, we denote its manual transcription sequence as:

$$
Y = \{y_i\}_{i=1}^{n},
$$

where $y_i$ denotes the manual reference transcription of the $i$-th segment. A candidate boundary $b$ divides the sequence into:

$$
G_b = \{y_i\}_{i=1}^{b-1},
$$

$$
R_b = \{y_i\}_{i=b}^{n}.
$$

Here, $G_b$ denotes the trusted prefix context before the boundary, and $R_b$ denotes the reference transcriptions corresponding to the target region after the boundary. The trusted prefix always starts from the beginning of the conversation and continuously extends to the segment immediately before the boundary. For each valid conversation, we retain only one final boundary, so that each conversation produces at most one RAMC-Corr sample. This avoids generating multiple overlapping samples from the same conversation.

\subsubsection{Candidate Boundary Generation}

We first enumerate candidate boundaries over the segment sequence of each conversation and apply preliminary filtering based on length and position constraints. Specifically, the trusted prefix before the boundary is required to contain a sufficient number of valid semantic segments and valid characters, so that it can provide usable historical context. At the same time, the trusted prefix must not exceed a predefined upper bound, in order to control the input size and reduce redundant history. The target region after the boundary is also required to satisfy minimum constraints on the number of segments and text length, ensuring that each sample contains a multi-segment correction target.

In addition, we constrain the relative position of the candidate boundary within the full conversation. This prevents the boundary from being placed too early, which would result in insufficient context, or too late, which would make the target region too short. Candidate boundaries that satisfy these constraints are then passed to the context continuity evaluation stage.

\subsubsection{Text Validity Filtering}

When computing length statistics and similarity scores, we only consider text segments with clear semantic content. Specifically, we filter out empty texts, noise markers, symbol-only segments, segments with an excessively high punctuation ratio, and short backchannel expressions commonly observed in real conversations.

This filtering is used only for statistics and similarity computation during boundary selection. It does not modify the original segment sequence in the final sample. The final samples still preserve the original conversation order, together with the corresponding speaker and timestamp information.

\subsubsection{Context Continuity Constraint}

To ensure that the trusted prefix and the target region remain semantically coherent, we compute a context continuity score for each candidate boundary. Specifically, we concatenate the valid texts in $G_b$ as the prefix text, and concatenate the valid texts from the first $K$ segments after the boundary as the target-head text $H_b$. We then compute the similarity between these two texts.

We use a weighted fusion of character n-gram similarity and semantic embedding similarity:

$$
s_b =
\alpha s_{\mathrm{ngram}}(G_b, H_b)
+
(1-\alpha)s_{\mathrm{emb}}(G_b, H_b),
$$

where $s_{\mathrm{ngram}}$ measures entity, keyword, and local lexical overlap, while $s_{\mathrm{emb}}$ measures topic-level semantic relatedness. The coefficient $\alpha$ is the fusion weight. We retain only candidate boundaries whose $s_b$ reaches the predefined threshold, and use this score as an important criterion for final boundary selection.

\subsubsection{Final Boundary Selection}

After filtering by length, position, and context continuity constraints, each conversation may retain multiple candidate boundaries. We further select one final boundary from these candidates to construct the corresponding RAMC-Corr sample. The selection process prioritizes context continuity across the boundary, while also preferring a trusted prefix with appropriate length and a reasonable boundary position. This ensures that each sample contains sufficient historical context while preserving enough subsequent content for correction.

For the candidate boundary $b$ of the $c$-th conversation, we define its selection score as:

$$
S_c(b)
=
w_s s_c(b)
-
w_l d_c(b)
-
w_r r_c(b).
$$

Here, $s_c(b)$ denotes the context continuity score across the boundary; $d_c(b)$ denotes the deviation between the trusted prefix length and the expected prefix length; $r_c(b)$ denotes the boundary position penalty, which is used to avoid boundaries that are too early or too late; and $w_s$, $w_l$, and $w_r$ are the corresponding weights. This scoring function favors split points with stronger context continuity, a moderately sized trusted prefix, and a reasonable boundary position.

The final boundary is defined as:

$$
b_c^{*}
=
\arg\max_{b \in \mathcal{B}_c}
S_c(b),
$$

where $\mathcal{B}_c$ denotes the set of candidate boundaries for the $c$-th conversation after length, position, and context continuity filtering. For each valid conversation, we retain only the boundary with the highest score. Therefore, each conversation produces at most one RAMC-Corr sample. If no candidate boundary satisfies the constraints, the conversation is excluded from the construction of RAMC-Corr. The specific values of each parameter are shown in the table \ref{tab:build_dataset_hyper}.
\begin{table}[t]
\centering
\setlength{\tabcolsep}{10pt}
\resizebox{0.95\columnwidth}{!}{%
\begin{tabular}{lll}
\hline
\textbf{Parameter} & \textbf{Value} & \textbf{Description} \\
\hline
embedding model & Qwen3-Embedding-8B & Embedding model \\
min prefix segments & 5 & Min. valid prefix segments \\
min prefix chars & 80 & Min. valid prefix chars \\
max prefix segments & 50 & Max. valid prefix segments \\
max prefix chars & 1200 & Max. valid prefix chars \\
min target segments & 10 & Min. valid target segments \\
min target chars & 150 & Min. valid target chars \\
target head \ensuremath{K} & 8 & Target-head segments \\
min similarity & 0.25 & Min. continuity score \\
min boundary ratio & 0.05 & Min. boundary position \\
max boundary ratio & 0.30 & Max. boundary position \\
target prefix segments & 20 & Expected prefix length \\
min boundary chars & 4 & Min. chars at boundary start \\
fusion weight \ensuremath{\alpha} & 0.4 & n-gram fusion weight \\
\ensuremath{w_s} & 0.70 & Continuity weight \\
\ensuremath{w_l} & 0.20 & Length penalty weight \\
\ensuremath{w_r} & 0.10 & Position penalty weight \\
\hline
\end{tabular}
}
\caption{Parameters used in the construction of RAMC-Corr. Min. and Max. denote minimum and maximum, respectively; chars denotes characters.}
\label{tab:build_dataset_hyper}
\end{table}




\subsection{Effect of in-context examples.}
\label{app:example_ablation}
We further study the effect of adding in-context examples to the correction prompt.
To construct the example pool, we first sampled 10 conversations from the training set and then sampled 200 utterances from these conversations.
From this pool, we manually selected and annotated 10 representative ASR errors as candidate in-context examples.

\begin{figure}[t]
    \centering
    \includegraphics[width=1.00\linewidth]{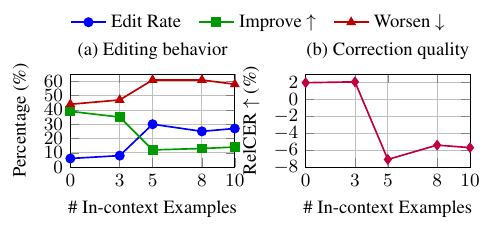}
    \caption{Effect of the number of in-context examples on editing behavior and correction quality.}
    \label{fig:examples_ablation}
\end{figure}

As shown in Figure~\ref{fig:examples_ablation}, simply increasing the number of examples does not consistently improve correction quality.
While using a small number of examples yields positive relative CER reduction, adding more examples makes the model edit more aggressively.
In particular, when the number of examples increases from 3 to 5, the edit rate rises substantially, but the proportion of worsened edits also increases, leading to a sharp drop in relative CER reduction.
This suggests that naively adding more static examples can induce over-correction, where the model edits more often but introduces more harmful changes.

These results indicate that the construction of in-context examples remains an important design factor.
More effective example selection, retrieval, and ordering strategies may be needed to ensure that demonstrations provide useful correction guidance rather than biasing the model toward unnecessary edits.

\subsection{Evidence Utilization Analysis}
\label{app:evidence_utilization}

\begin{table}[t]
\centering
\setlength{\tabcolsep}{10pt}
\resizebox{0.95\columnwidth}{!}{%
\small
\begin{tabular}{llrrrr}
\toprule
Setting & Retrieval & Count & Edit & Improve  $\uparrow$ & Worsen  $\downarrow$ \\
\midrule
Zero-shot & With & 13,304 & 6.16 & 39.19 & 42.98 \\
Zero-shot & Without & 402 & 9.20 & 35.14 & 54.05 \\
Few-shot & With & 13,106 & 7.55 & 35.89 & 45.60 \\
Few-shot & Without & 600 & 10.17 & 26.23 & 59.02 \\
\bottomrule
\end{tabular}
}
\caption{Edit outcomes grouped by retrieval availability for Qwen3.5-4B. Edit denotes Edit Rate, while Improve and Worsen denote Improve@Edit and Worsen@Edit. All rates are percentages.}
\label{tab:retrieval_availability}
\end{table}

We compute all statistics at the correction-record level.
Retrieval availability indicates whether at least one ontology entry is retrieved and provided to the correction model.
A record is counted as edited if the corrected output differs from the ASR hypothesis.
Improve@Edit and Worsen@Edit are computed among edited records, measuring the proportions of edits that decrease or increase CER, respectively.
Memory updates are detected when the ontology working memory is modified after processing the current record.

Table~\ref{tab:retrieval_availability} analyzes edit outcomes by retrieval availability for Qwen3.5-4B.
Retrieved ontology evidence is available for 97.07\% and 95.62\% of records under zero-shot and few-shot prompting, respectively, with memory updates detected in 6.48\% and 6.68\% of records.
Across both settings, records with retrieved evidence show lower Edit Rate and Worsen@Edit than those without retrieval, suggesting more conservative and less harmful editing.

\section{Prompt Templates}
\label{app:prompt_templates}
This section provides the prompt templates used in our experiments.
Table~\ref{tab:prompt_no_wm} shows the correction prompt without ontology working memory, Table~\ref{tab:prompt_wm_correction} shows the memory-augmented correction prompt, and Table~\ref{tab:ontology_instruction} shows the prompt for extracting candidate ontology entries.

\definecolor{process_text}{RGB}{180,0,3}
\definecolor{agent_text}{RGB}{150,10,200}
\definecolor{llm_text}{RGB}{55,180,9}

\begin{table}[t]
    \centering

\begin{tikzpicture}

    \node[inner sep=0pt] (table) {
        \begin{tabular}{p{0.45\textwidth}}
            \rowcolor{gray!15}
            \textbf{ASR Correction without Ontology Memory}
            \\
            \hline

            \textbf{User}: You are an ASR text correction expert. Your task is to identify words in the current transcription that need correction based on the current ASR text and recent dialogue context, and output the corrected text.
            \\

            \\
            \textbf{Task priorities:}
            \\
            1. Prioritize ASR correction and make the corrected text as accurate as possible.
            \\
            2. Only correct ASR recognition errors. Do not polish, summarize, expand, or rewrite the style.
            \\
            3. If there is no clear error, keep the corrected text identical to the current transcription.
            \\

            \\
            \textbf{Correction principles:}
            \\
            - First inspect the current sentence, then refer to the historical context.
            \\
            - Focus on keywords, named entities, terms, numbers, collocations, and contextually inconsistent words.
            \\
            - Do not change the topic based on a single abrupt word unless the context clearly indicates a topic shift.
            \\
            - If the context does not provide enough evidence, keep the original word.
            \\
            - Preserve the original word order, spoken style, and amount of information as much as possible.
            \\

            \\
            \textbf{Forbidden behaviors:}
            \\
            - Do not turn correction into polishing, summarization, or rewriting.
            \\
            - Do not add information that is not expressed in the current sentence.
            \\
            - Do not delete semantically meaningful spoken components.
            \\
            - Do not output explanatory completion text.
            \\

            \\
            \textbf{Recent dialogue context:}
            \\
            \textcolor{process_text}{\{recent\_corrected\_context\}}
            \\

            \\
            \textbf{Current ASR transcription:}
            \\
            \textcolor{process_text}{\{current\_asr\_text\}}
            \\

            \\
            Please output only the corrected text.
            \\

            \\
            \textbf{LLM}: \textcolor{llm_text}{\{corrected\_text\}}
        \end{tabular}
    };

    \draw[rounded corners=5pt, line width=1pt]
        (table.north west) -- (table.north east) --
        (table.south east) -- (table.south west) -- cycle;

\end{tikzpicture}
\caption{Prompt template for ASR correction without ontology memory.}
\label{tab:prompt_no_wm}
\end{table}
\definecolor{process_text}{RGB}{180,0,3}
\definecolor{agent_text}{RGB}{150,10,200}
\definecolor{llm_text}{RGB}{55,180,9}

\begin{table}[t]
    \centering
    \footnotesize
    \setlength{\tabcolsep}{1.2mm}

\begin{tikzpicture}

    \node[inner sep=0pt] (table) {
        \begin{tabular}{p{0.46\textwidth}}
            \rowcolor{gray!15}
            \textbf{ASR Correction with Ontology Memory}
            \\
            \hline

            \textbf{User}: You are an ASR text correction expert. Correct the current transcription using recent dialogue context, active entities, and retrieved ontology working memory (WM).
            \\

            \\
            \textbf{Task priorities:}
            \\
            \dots
            \\

            \\
            \textbf{Correction principles:}
            \\
            \dots
            \\
            - Inspect the current sentence first, then recent context, and finally WM.
            \\
            - If a word conflicts with recent topic, active entities, or WM, correct it only after confirming it is ASR-suspicious.
            \\
            - If neither context nor WM provides enough evidence, keep the original word.
            \\
            - Recent contextual evidence has priority over distant WM.
            \\

            \\
            \textbf{WM usage rules:}
            \\
            - WM is auxiliary evidence, not a forced replacement basis.
            \\
            - Retrieved knowledge may contain noise.
            \\
            - Do not replace text with a WM entity solely because it appears in WM.
            \\
            - Use WM only when the ASR text is suspicious or contextually inconsistent.
            \\
            - If WM conflicts with the current sentence and no clear ASR error exists, trust the current sentence.
            \\

            \\
            \textbf{Forbidden behaviors:}
            \\
            \dots
            \\
            - Do not force the sentence to match WM or change the topic for WM.
            \\
            - Do not change ordinary words into WM entities or terms without evidence.
            \\

            \\
            \textbf{Inputs:}
            \\
            Retrieved WM: \textcolor{agent_text}{\{retrieved\_knowledge\}}
            \\
            Active entities: \textcolor{agent_text}{\{active\_mentions\}}
            \\
            Recent context: \textcolor{process_text}{\{recent\_corrected\_context\}}
            \\
            Current ASR: \textcolor{process_text}{\{current\_asr\_text\}}
            \\

            \\
            \textbf{LLM}: \textcolor{llm_text}{\{corrected\_text\}}
        \end{tabular}
    };

    \draw[rounded corners=5pt, line width=1pt]
        (table.north west) -- (table.north east) --
        (table.south east) --
        (table.south west) -- cycle;

\end{tikzpicture}
\caption{Prompt template for ASR post-correction with retrieved ontology memory and recent active entities. Shared instructions inherited from the no-memory prompt are omitted for brevity, while WM-specific constraints are shown.}
\label{tab:prompt_wm_correction}
\end{table}
\begin{table}[t]
    \centering

\begin{tikzpicture}

    \node[inner sep=0pt] (table) {
        \begin{tabular}{p{0.45\textwidth}}
            \rowcolor{gray!15}
            \textbf{Ontology Instruction}
            \\
            \hline

            \textbf{Role Definition}
            \\
            The model acts as a working-memory update expert for an ASR correction system.
            Its task is to extract stable nodes that can be safely written into working memory (WM)
            based on the original ASR transcript \texttt{transcript\_text} and the corrected text
            \texttt{corrected\_text}. These nodes should be clearly useful for subsequent ASR correction.
            The output must be strictly valid JSON.
            \\
            \textbf{Core Objective}
            \\
            This task is not intended to build a complete knowledge graph, nor to summarize the text content.
            It should only record anchors that explicitly appear in \texttt{corrected\_text}, are stable and reusable,
            and are clearly helpful for future ASR correction.
            \\

            \\
            \textbf{A valid node must satisfy:}
            \\
            1. The \texttt{canonical\_name} must come from a phrase explicitly appearing in
            \texttt{corrected\_text}.
            \\
            2. It must be a stable and reusable entity, term, special identifier, special version,
            code, abbreviation, proprietary expression, work title, organization name, role name,
            group name, product name, domain term, core concept, or key numerical value.
            \\
            3. If it is misrecognized, incorrectly segmented, omitted, inserted, or mistranscribed by ASR
            in the future, it should significantly affect the semantics or character error rate (CER).
            \\
            4. If there is a corresponding erroneous fragment in \texttt{transcript\_text}, it should be
            recorded as an \texttt{asr\_noise\_form}.
            \\
            5. It must not be an ordinary word, ordinary phrase, sentence-level rewrite, semantic summary,
            polishing completion, or temporary description.
            \\
            \textbf{Recent context:} \textcolor{process_text}{\{history/context\}}
            \\

            \textbf{Original ASR transcription:} \textcolor{process_text}{\{transcript\_text\}}
            \\

            \textbf{Corrected text:} \textcolor{process_text}{\{corrected\_text\}}
            \\

            \textbf{Output format:}
            \\
            Please output only valid JSON with the field \texttt{ontology\_nodes}.
            If no valid node exists, output an empty list.
            \\

        \end{tabular}
    };

    \draw[rounded corners=5pt, line width=1pt]
        (table.north west) -- (table.north east) --
        (table.south east) -- (table.south west) -- cycle;

\end{tikzpicture}
\caption{Prompt template for ontology extraction.}
\label{tab:ontology_instruction}
\end{table}

\end{CJK}
\end{document}